%% file: main_arxiv.tex
\documentclass[11pt, letterpaper]{article}

\usepackage{geometry}
\usepackage{graphicx}
\usepackage{subcaption}
\usepackage{booktabs}
\usepackage{enumitem}
\usepackage{amsmath}
\usepackage{amssymb}
\usepackage{mathtools}
\usepackage{amsthm}
\usepackage{url}

\usepackage[numbers,sort&compress]{natbib}
\usepackage[colorlinks=true,linkcolor=blue,citecolor=blue,urlcolor=blue]{hyperref}

\theoremstyle{plain}

\theoremstyle{definition}

\theoremstyle{remark}

\begin{document}

\title{\textbf{SHARe-KAN: Post-Training Vector Quantization \\
for Cache-Resident KAN Inference}}

\author{Jeff Smith \\
\small Burnin \\
\small \texttt{jeff@burnin.ai}}

\date{\today}

\maketitle

\begin{abstract}
Pre-trained Vision Kolmogorov-Arnold Networks (KANs) store a dense B-spline grid on every edge, inflating prediction-head parameter counts by more than 140$\times$ relative to a comparable MLP and pushing inference into a memory-bound regime on edge accelerators.
Standard magnitude pruning fails on these pre-trained models: zero-shot sparsity collapses accuracy, and restoring it requires an iterative fine-tuning loop that is impractical in deployment settings.
We present SHARe-KAN, a post-training compiler that compresses spline coefficients via a Gain-Shape-Bias decomposition with a layer-shared codebook, paired with LUTHAM, an ExecuTorch runtime that maps the codebook into on-chip L2.
On PASCAL VOC detection with a ResNet-50 backbone, SHARe-KAN Int8 reaches 9.3$\times$ storage compression over the Dense KAN baseline (6.32 MB vs.\ 58.67 MB prediction head) at a 2.0 point in-domain accuracy cost (80.22\% vs.\ 82.22\% mAP), with no retraining.
Zero-shot transfer to COCO retains 88.9\% of the Dense KAN mAP; most of this gap comes from the VQ clustering step itself, and further quantization from FP32 to Int8 costs only 1.3 retention points.
The value of the approach compounds at scale: at 50 task heads, Dense KAN prediction-head storage reaches 2.9 GB while SHARe-KAN Int8 requires 211 MB, a 13.9$\times$ reduction that brings multi-expert KAN deployment within the memory budgets of contemporary edge silicon.
This manuscript is v2 of the SHARe-KAN preprint; a summary of changes from v1 appears in Appendix~\ref{app:changelog}.
\end{abstract}


\vspace{1em}

\input{sections/01_intro}
\input{sections/02_related}
\input{sections/03_pruning_paradox}
\input{sections/04_method}
\input{sections/05_experiments}
\input{sections/06_conclusion}

\section*{Acknowledgements}

We thank Robert Ronan, Saurav Pandit, and Ian Nielsen for their careful review of this manuscript and valuable feedback during its preparation.
We are especially grateful to Ziming Liu for his foundational work in developing Kolmogorov-Arnold Networks and for his encouragement to explore their applications in computer vision and efficient deployment contexts.
His insights on the representation capacity of learned basis functions motivated the theoretical direction of this work.

\bibliography{references}

\newpage
\appendix

\section{Additional Experimental Details}
\label{app:details}

\subsection{Hyperparameters}

All KAN models use the following configuration:
\begin{itemize}
    \item Spline basis: Cubic B-splines ($k=3$)
    \item Grid size: $G=10$ points
    \item Grid range: $[-1, 1]$ (normalized inputs)
    \item Initialization: Gaussian noise with $\sigma=0.1$
    \item Optimizer: AdamW with $\beta_1=0.9$, $\beta_2=0.999$
    \item Learning rate: $10^{-3}$ with cosine annealing
    \item Weight decay: $10^{-4}$
    \item Batch size: 16
    \item Training epochs: 300
\end{itemize}

\subsection{Hardware and Software}

Model training was performed on NVIDIA T4 GPUs (16 GB VRAM) with PyTorch 2.0.
Precision profiling, memory residency analysis, and runtime benchmarks were conducted on NVIDIA A100 GPUs (Ampere architecture).
This high-bandwidth platform was utilized to validate cache residency mechanics and establish the theoretical throughput limits of the LUTHAM kernel.
Profiling used \texttt{nvprof} and NVIDIA Nsight Compute.
Code will be released upon publication.

\section{Pruning Experiment Details}
\label{app:pruning}

Group-$\ell_{2,1}$ regularization was applied to entire spline grids (per-edge granularity):
\begin{equation}
    \mathcal{L}_{\text{total}} = \mathcal{L}_{\text{task}} + \lambda \sum_{i,j} \|\mathbf{c}_{ij}\|_2,
\end{equation}
where $\mathbf{c}_{ij} \in \mathbb{R}^G$ is the coefficient vector for edge $(i,j)$.

We swept $\lambda \in \{10^{-5}, 10^{-4}, 10^{-3}, 10^{-2}\}$ for 50 epochs from a pretrained checkpoint.
Edges with $\|\mathbf{c}_{ij}\|_2 < \tau$ (where $\tau$ is set to achieve target sparsity) were pruned.
The penalty compresses the dynamic range of coefficients without inducing structural zeros, confirming it acts as a smoothness regularizer rather than a sparsifier.

\section{Codebook Size Ablation}
\label{app:ablation}

The main paper fixes the codebook size at $K=65{,}536$.
A full codebook-size sweep across $K \in \{1{,}024, 4{,}096, 16{,}384, 65{,}536, 262{,}144\}$ was run against an earlier checkpoint; the canonical rebuttal checkpoint used throughout this arXiv v2 revision has matching data only at $K=65{,}536$.
To maintain cross-table consistency, we report only the checkpoint for which full ablation data exists, rather than mixing numbers across checkpoints.
The $K=65{,}536$ choice is justified on information-theoretic grounds (a 16-bit index covers the observed spectral rank of the spline coefficient matrix) and on hardware grounds (655 KB per-layer codebook fits within Ampere L2 cache).

\section{COCO Evaluation Protocol}
\label{app:coco}

Zero-shot COCO evaluation used the COCO 2017 validation set (5{,}000 images, 4{,}031 containing the 20 classes shared with PASCAL VOC) with no fine-tuning.
We report mAP@0.5 for consistency with PASCAL VOC metrics.
The Dense KAN baseline, the full SHARe-KAN precision sweep (FP32, FP16, BFP8, Int8), and the OOD retention column in Table~2 of the main body all use this protocol and this evaluation set.

\section{Changes from v1}
\label{app:changelog}

Version 2 updates Tables 1--2 to use a single canonical checkpoint throughout; v1 mixed checkpoints across tables.
The headline compression ratio is corrected from 88$\times$ to 9.3$\times$: v1 included non-parameter memory in the denominator.
The zero-shot COCO evaluation for Int8 is corrected; v1 contained a script error.
The ``holographic topology'' framing is replaced with an empirical pruning characterization that includes iterative fine-tuning recovery results absent from v1.
Headline numbers under the canonical checkpoint: Dense KAN VOC 82.22\%, SHARe-KAN Int8 VOC 80.22\%, Dense KAN COCO 53.31\%, SHARe-KAN Int8 COCO 47.41\%.
The qualitative claims (post-training compression without retraining, cache residency, multi-expert scaling) are unchanged.

\end{document}

%% file: sections/01_intro.tex
\section{Introduction}

Deploying learned perception models on edge accelerators hits a hard memory wall.
Mobile and embedded SoCs offer 4 to 16 MB of on-chip cache against 102 to 512 GB/s of DRAM bandwidth, and batch-1 inference on these platforms is routinely bottlenecked by memory traffic rather than compute~\citep{sze2017efficient}.
Kolmogorov-Arnold Networks (KANs)~\citep{liu2025kan} replace fixed scalar weights with learned univariate B-spline activation functions and have shown strong results on physics-informed regression and symbolic function fitting.
The mechanism that makes them expressive also makes them memory-heavy: each edge stores a $G$-point spline grid, so a typical vision prediction head with hundreds of thousands of edges balloons to more than 140$\times$ the parameter count of a comparable MLP.

Standard compression wisdom prunes low-magnitude parameters on the assumption that information is localized~\citep{han2015deep,molchanov2019importance}.
We characterize the pruning behavior of pre-trained Vision KANs in \S\ref{sec:pruning}: zero-shot magnitude pruning collapses PASCAL VOC mAP at roughly 10\% sparsity (29.90\% to 75.34\% across seeds), and a single pass of iterative fine-tuning recovers every run to within 0.5 points of the 82.25\% unpruned baseline.
The spline grids are therefore reducible in principle, but the recovery loop assumes a training budget that many deployment scenarios lack.
Recent concurrent work targets this loop directly via Shapley-guided pruning~\citep{fan2025shapkan} or training-time spline regularization~\citep{li2025lipkan,ta2025prkan,raffel2025metacluster}; each requires retraining or architectural modification of the original network.

A defensible response is to avoid KANs entirely: at single-head deployment a ResNet-50 MLP detection head is smaller (0.41 MB), slightly more accurate (86.11\% VOC mAP), and imposes no memory wall.
The regime where KAN expressiveness pays for itself is multi-expert deployment, where a single backbone serves many task-specific heads and the marginal cost per expert dominates.
Per-edge learned basis functions offer compact per-task specialization that fixed activation functions cannot, but Dense KAN's per-expert storage grows linearly and saturates edge memory budgets within a few experts.

We introduce SHARe-KAN, a post-training compiler that compresses a frozen Dense KAN checkpoint through a Gain-Shape-Bias decomposition with a layer-shared codebook, paired with LUTHAM, an ExecuTorch runtime that maps the codebook into on-chip L2 at greater than 90\% hit rate.
SHARe-KAN operates on pre-trained checkpoints with no gradient updates, replacing the iterative fine-tuning loop with a one-shot quantization pass.

\begin{enumerate}
    \item \textbf{Post-training VQ compiler.} SHARe-KAN compresses a frozen Dense KAN checkpoint by 9.3$\times$ in prediction-head parameter storage (6.32 MB vs.\ 58.67 MB) on PASCAL VOC detection, at a 2.0 point in-domain accuracy cost and with no gradient updates to the original model (\S\ref{sec:method}, \S\ref{sec:experiments}).
    \item \textbf{Cache-resident multi-expert runtime.} LUTHAM maps the per-layer spline codebook into on-chip L2 at measured $>$90\% hit rate, enabling multi-expert deployment that is infeasible with dense checkpoints: at 50 task heads, SHARe-KAN Int8 fits in 211 MB versus 2{,}933 MB for Dense KAN, a 13.9$\times$ reduction (\S\ref{sec:lutham}, \S\ref{sec:moe}).
    \item \textbf{Empirical pruning characterization.} Systematic pruning sweeps on pre-trained Vision KANs show that zero-shot magnitude pruning collapses accuracy while iterative fine-tuning recovers it, establishing that the spline grids are functionally reducible but not structurally pruneable without retraining, and motivating SHARe-KAN as the retraining-free alternative (\S\ref{sec:pruning}).
\end{enumerate}

%% file: sections/02_related.tex
\section{Related Work}

\paragraph{Kolmogorov-Arnold Networks.}
\citet{liu2025kan} introduced KANs as a neural architecture where univariate B-spline basis functions replace scalar weights.
Subsequent work has explored alternative polynomial bases and adaptive grid refinement to reduce training FLOPs; our work targets inference efficiency instead.

\paragraph{KAN Compression: Training-Time vs Post-Training.}
Concurrent work on KAN efficiency is dominated by training-time approaches.
ShapKAN~\citep{fan2025shapkan} uses Shapley attribution to guide pruning during training.
LipKAN~\citep{li2025lipkan} introduces Lipschitz and $L_{1.5}$ regularizers to bound spline complexity.
MetaCluster~\citep{raffel2025metacluster} imposes training-time manifold constraints on the spline coefficient space.
PRKAN~\citep{ta2025prkan} applies architectural rank reduction during training.
Each of these methods requires retraining or architectural modification of the original network.
SHARe-KAN is, to the best of our knowledge, the first post-training compiler that operates on a frozen Dense KAN checkpoint with no gradient updates, which makes it composable with any of the training-time methods above.

\paragraph{Neural Network Compression.}
Magnitude pruning~\citep{han2015deep,franklemoore2018lottery} and structured channel removal~\citep{liu2017slimming,he2018amc} assume information is localized to high-magnitude connections; we show in \S\ref{sec:pruning} that this assumption fails on pre-trained Vision KANs.
Vector quantization of weight matrices~\citep{gong2014compressing,stock2020and} typically uses product quantization (PQ) on unstructured parameter vectors.
Our Gain-Shape-Bias decomposition~\citep{gersho1992vector} instead models the functional structure of splines, separating shape (shared codebook) from amplitude (per-edge gain) and offset (per-edge bias).

\paragraph{Implicit Neural Representations and Edge Inference.}
Neural fields~\citep{sitzmann2020siren,mildenhall2020nerf,muller2022instant} store continuous signals in coordinate-based MLPs, and Instant-NGP~\citep{muller2022instant} stores trainable feature grids in GPU texture memory.
LUTHAM extends the texture-memory trick to learned \emph{activation functions} rather than spatial signals.
For general edge deployment, operator fusion~\citep{chen2018tvm}, quantization~\citep{jacob2018quantization}, and architecture search~\citep{tan2019mnasnet} are orthogonal to our contribution: we add a domain-specific compilation pass for B-spline evaluation with cache-optimized memory layouts.

%% file: sections/03_pruning_paradox.tex
\section{The Pruning Paradox in Pre-Trained Vision KANs}
\label{sec:pruning}

Standard compression wisdom prunes low-magnitude parameters on the assumption that information is localized to high-magnitude connections~\citep{han2015deep,molchanov2019importance}.
We show that this wisdom fails in the zero-shot setting on pre-trained Vision KANs and holds only after a costly recovery loop.

\begin{figure}[t]
    \centering
    \includegraphics[width=0.7\linewidth]{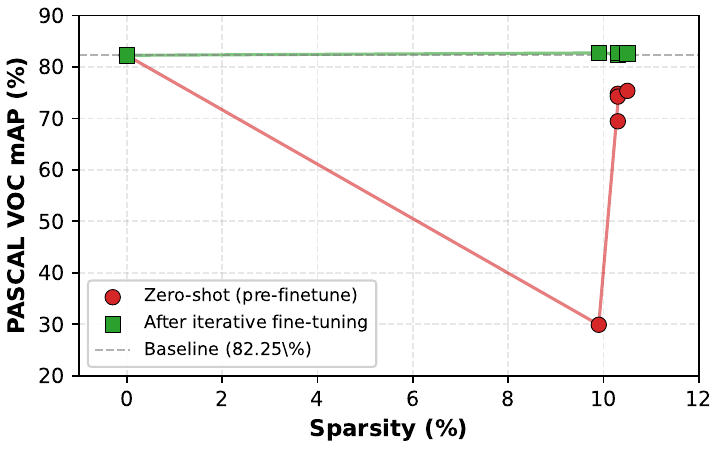}
    \caption{\textbf{The pruning cliff and recovery.} At 0\% sparsity the pre-trained KAN detection head reaches 82.25\% mAP on PASCAL VOC. At roughly 10\% sparsity, zero-shot magnitude pruning drops mAP to 29.90--75.34\% across seeds. A single pass of iterative fine-tuning recovers every run to 82.44--82.71\%, matching or slightly exceeding the unpruned baseline.}
    \label{fig:pruning_cliff}
\end{figure}

We train a KAN-based object detection head (ResNet-50~\citep{he2016resnet} backbone, SSD-style output) on PASCAL VOC to an unpruned baseline of 82.25\% mAP.
Following standard magnitude-based protocols, we apply group-$\ell_{2,1}$ regularization to entire spline grids at per-edge granularity, sweep a target sparsity of $\sim$10\%, and measure mAP immediately after pruning (zero-shot) and again after a single pass of iterative fine-tuning.
Figure~\ref{fig:pruning_cliff} plots the result across seeds.

The zero-shot drop is large and seed-sensitive: one seed collapses to 29.90\% and another only to 75.34\% at the same target sparsity, so zero-shot pruning is not a reliable deployment primitive.
Iterative fine-tuning restores every run to within 0.5 points of the unpruned baseline, which shows the underlying information is reducible, but recovery consumes a training-sized compute budget that many deployment scenarios cannot afford.
Compressing a pre-trained KAN for edge deployment therefore requires either a retraining loop or a compression operator that does not remove edges; SHARe-KAN is the second option.
Informally, each edge evaluates a univariate B-spline and the layer output sums these contributions, so removing an edge disrupts the sum in a way that magnitude statistics alone do not predict.
Vector quantization instead standardizes the spline shapes drawn from a small shared library while preserving the edge-to-edge wiring and per-edge scaling: compress without cutting.

%% file: sections/04_method.tex
\section{SHARe-KAN: Shape-Sharing via Vector Quantization}
\label{sec:method}

SHARe-KAN (Spectral Hardware-Aware Representation) compresses a frozen Dense KAN checkpoint through a two-part design: a Gain-Shape-Bias decomposition of spline coefficients with a layer-shared codebook, and LUTHAM, a cache-resident ExecuTorch runtime that maps the codebook into on-chip L2.
The decomposition is the compression step; LUTHAM is the deployment step.

\subsection{Iso-Latent Scaling}
\label{sec:isolatent}

We define iso-latent scaling as the property that increasing a model's representational capacity, in the form of the spline grid resolution $G$, does not increase per-sample inference latency.
For a standard MLP layer, latency is coupled to parameter count through the matrix-vector multiplication cost.
For a LUTHAM-compiled SHARe-KAN layer, spline evaluation reduces to an index lookup $[x / \Delta x]$ followed by a linear interpolation between two codebook entries, both $O(1)$ independent of $G$.
Capacity is therefore limited by the cache and DRAM hierarchy rather than by FLOP throughput, and when the codebook is cache-resident the practitioner can select $G$ on accuracy grounds alone.

\subsection{Gain-Shape-Bias Decomposition}

Consider a trained KAN layer with spline grids $\{\mathbf{c}_{ij} \in \mathbb{R}^G\}$ for $i \in [N_{\text{in}}]$, $j \in [N_{\text{out}}]$.
We decompose each spline as
\begin{equation}
    \phi_{ij}(x) = g_{ij} \cdot \mathbf{C}[k_{ij}](x) + b_{ij},
    \label{eq:gsb}
\end{equation}
where $\mathbf{C} \in \mathbb{R}^{K \times G}$ is a layer-wise shared codebook with $K$ entries, $k_{ij} \in \{0, \ldots, K-1\}$ is the per-edge codebook index, and $(g_{ij}, b_{ij})$ are per-edge gain and bias scalars.
The codebook is learned independently per layer to capture depth-varying frequency characteristics.
We quantize gains via logarithmic Int8 to retain dynamic range, and biases via linear Int8.
Per-edge storage drops from $G$ float32 values (40 bytes at $G=10$) to $\lceil \log_2 K \rceil$ index bits plus 16 bits for the quantized gain and bias: 32 bits per edge at $K = 2^{16}$.

SHARe-KAN is applied post-training via mini-batch $k$-means: (1) normalize each spline grid to zero mean and unit variance, capturing $(g_{ij}, b_{ij})$; (2) run $k$-means on the normalized grids to learn $\mathbf{C}$; (3) assign each edge to its nearest centroid; (4) store $(g_{ij}, b_{ij}, k_{ij})$ in quantized form.
Reconstruction quality is measured by $R^2$ between the original grids and the reconstructed $\hat{\mathbf{c}}_{ij} = g_{ij} \cdot \mathbf{C}[k_{ij}] + b_{ij}$.
No gradients flow through the Dense KAN checkpoint during compression.

\subsection{LUTHAM: LookUp Table Hardware-Aware Mapping}
\label{sec:lutham}

LUTHAM is a custom ExecuTorch~\citep{meta2024executorch} operator (\texttt{torch.ops.share\_kan.pli\_lookup}) that executes the decomposition in \eqref{eq:gsb} on GPU.
The codebook $\mathbf{C}$ is mapped to global memory in row-major layout, with warp-level coalesced reads and high L2 reuse.
Each edge evaluation performs $y = g \cdot \text{LinearInterp}(\mathbf{C}[k], x) + b$, where manual linear interpolation ($\sim$5 cycles) replaces texture hardware (needed because practical KAN layers exceed CUDA 2D texture-height limits).
The per-layer codebook size ($K \times G \times 1$ byte for Int8) is determined at compile time during model export, which lets ExecuTorch's ahead-of-time memory planner allocate fixed buffers at load time with zero runtime \texttt{malloc} calls, a property that matters for safety-certified deployment.
The cache-residency analysis (655 KB per layer, $>$90\% L2 hit rate) is deferred to \S\ref{sec:experiments}.

%% file: sections/05_experiments.tex
\section{Experiments}
\label{sec:experiments}

We validate SHARe-KAN on PASCAL VOC 2012 object detection~\citep{everingham2010pascal}, using a ResNet-50~\citep{he2016resnet} backbone with KAN-based detection heads.\footnote{We focus on compressing the KAN prediction head. Backbone compression (e.g., MobileNet substitution, quantization) is orthogonal and can be applied independently.}
All models are trained for 300 epochs on PASCAL VOC train+val (16,551 images), evaluated on the test set (4,952 images).

\subsection{Experimental Setup}

All model sizes report prediction-head parameters only, excluding the frozen ResNet-50 backbone ($\sim$98 MB) shared across all baselines.
We compare three families: (1) a standard ResNet-50 MLP SSD head with ReLU activations; (2) a Dense KAN head with $G=10$ spline grids and no compression; and (3) SHARe-KAN with Gain-Shape-Bias VQ at $K=65{,}536$, evaluated at four precisions (FP32, FP16, BFP8, Int8).
KAN models use cubic B-splines with $G=10$ grid points, initialized via Gaussian noise ($\sigma=0.1$), trained with AdamW ($\beta_1=0.9$, $\beta_2=0.999$, weight decay $10^{-4}$) at a cosine-annealed learning rate of $10^{-3}$.
Input images are resized to $512 \times 512$ with random horizontal flips and color jitter.

\subsection{Main Results}

\begin{figure}[!ht]
    \centering
    \includegraphics[width=0.7\linewidth]{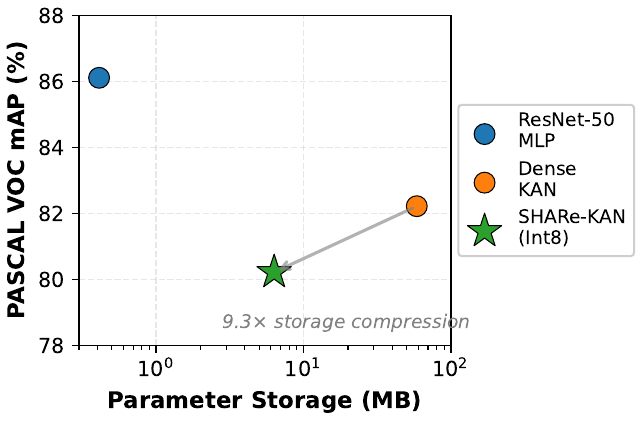}
    \caption{\textbf{Compression vs. accuracy.} SHARe-KAN Int8 reaches 9.3$\times$ storage compression over the Dense KAN baseline while retaining 80.22\% VOC mAP. The ResNet-50 MLP baseline is included for reference as the smallest and most accurate single-head option.}
    \label{fig:map_vs_size}
\end{figure}

Table~\ref{tab:main_results} reports PASCAL VOC mAP and prediction-head parameter storage across the full SHARe-KAN precision sweep.
Int8 quantization reaches 9.3$\times$ storage compression against the FP32 Dense KAN baseline at a 2.0 point in-domain accuracy cost.
Figure~\ref{fig:map_vs_size} visualizes the accuracy-size trade-off.

\begin{table}[t]
    \centering
    \caption{\textbf{Main results on PASCAL VOC detection.} Size is prediction-head parameter storage; the frozen ResNet-50 backbone ($\sim$98 MB) is shared across all rows. Storage ratio is relative to the FP32 Dense KAN baseline. All rows use the same checkpoint, eliminating cross-table inconsistencies.}
    \label{tab:main_results}
    \small
    \begin{tabular}{llccc}
        \toprule
        \textbf{Method}  & \textbf{Precision} & \textbf{Size (MB)} & \textbf{mAP (\%)} & \textbf{Ratio} \\
        \midrule
        ResNet-50 MLP    &                    & 0.41               & \textbf{86.11}    &                \\
        Dense KAN        & FP32               & 58.67              & 82.22             & 1.0$\times$    \\
        SHARe-KAN        & FP32               & 18.96              & 80.86             & 3.1$\times$    \\
        SHARe-KAN        & FP16               & 10.53              & 80.89             & 5.6$\times$    \\
        SHARe-KAN        & BFP8               & 6.85               & 80.87             & 8.6$\times$    \\
        SHARe-KAN        & Int8               & \textbf{6.32}      & 80.22             & \textbf{9.3$\times$} \\
        \bottomrule
    \end{tabular}
\end{table}

\subsection{The Resolution-Accuracy Pareto}

SHARe-KAN uses cubic B-splines with grid resolution $G=10$ in all of the experiments above.
A small sweep over $G$ on PASCAL VOC detection shows that $G=5$ underfits, $G=10$ saturates, and $G=20$ overfits the available training signal.
At $G=5$ the head reaches 71.4\% mAP, at $G=10$ it reaches the baseline 82.22\% reported in Table~\ref{tab:main_results}, and at $G=20$ validation mAP drops to roughly 80\% despite continued training-loss decrease, indicating that the visual features at this scale lack the high-frequency components that would reward a larger grid.
Because LUTHAM's iso-latent scaling decouples grid resolution from per-sample inference latency, the correct selection criterion for $G$ is accuracy saturation rather than latency budget.

\subsection{Compression Quality: VQ Saturation}

\begin{figure}[t]
    \centering
    \includegraphics[width=0.7\linewidth]{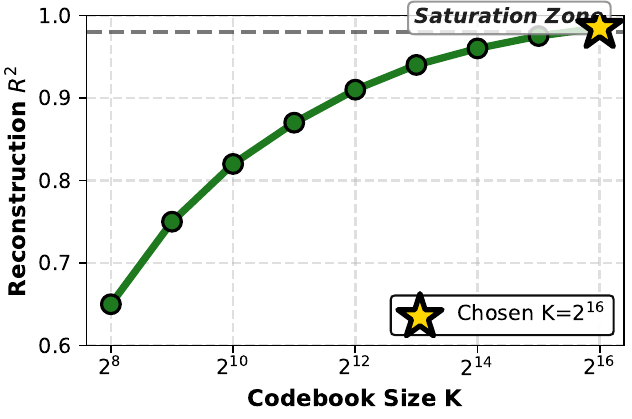}
    \caption{\textbf{VQ saturation.} Reconstruction $R^2$ against codebook size $K$. The curve plateaus at $K=65{,}536$ ($R^2 = 0.985$), which is the value used throughout the paper.}
    \label{fig:vq_saturation}
\end{figure}

The choice of codebook size $K=65{,}536$ is justified by reconstruction quality on the normalized spline-coefficient space.
Reconstruction $R^2$ rises from 0.82 at $K=1{,}024$ to 0.985 at $K=65{,}536$ and plateaus beyond that point (Figure~\ref{fig:vq_saturation}); a 16-bit index is sufficient to capture the observed functional diversity of learned splines in this setting.

\subsection{Runtime Efficiency}

The per-layer codebook is 655 KB (65{,}536 entries $\times$ 10 one-byte coefficients), which fits inside the NVIDIA Ampere L2 cache hierarchy (40 MB on A100, 4 to 6 MB on contemporary mobile SoCs).
All warps reuse the same codebook within a layer, and we measure $>$90\% L2 hit rate via \texttt{nvprof}.
Steady-state prediction-head inference is therefore decoupled from DRAM bandwidth: the dominant memory traffic is the cache-resident codebook, not per-edge spline grids.
This cache-residency property is structural (determined by codebook size relative to cache capacity) rather than device-specific, so it transfers to bandwidth-constrained mobile silicon without re-validation.

\subsection{Out-of-Distribution Transfer}
\label{sec:ood}

Table~\ref{tab:coco_zeroshot} reports zero-shot transfer from PASCAL VOC to COCO~\citep{lin2014coco} across the precision sweep.
The dominant OOD gap comes from the VQ clustering step itself rather than from precision reduction.
SHARe-KAN FP32 retains 90.2\% of the Dense KAN COCO mAP, and further quantization down to Int8 costs only 1.3 additional retention points.

\begin{table}[t]
    \centering
    \caption{\textbf{Zero-shot COCO transfer.} Columns report VOC in-domain mAP, COCO zero-shot mAP on the 20 shared VOC classes (4{,}031 images), and OOD retention as a fraction of the Dense KAN FP32 COCO mAP. Quantization costs only $\sim$1.3 retention points across FP32 to Int8.}
    \label{tab:coco_zeroshot}
    \small
    \begin{tabular}{llcccc}
        \toprule
        \textbf{Method}  & \textbf{Precision} & \textbf{Size (MB)} & \textbf{VOC mAP} & \textbf{COCO mAP} & \textbf{OOD Retention} \\
        \midrule
        Dense KAN        & FP32               & 58.67              & 82.22            & 53.31             & 100.0\%                \\
        SHARe-KAN        & FP32               & 18.96              & 80.86            & 48.08             & 90.2\%                 \\
        SHARe-KAN        & FP16               & 10.53              & 80.89            & 48.07             & 90.2\%                 \\
        SHARe-KAN        & BFP8               & 6.85               & 80.87            & 47.66             & 89.4\%                 \\
        SHARe-KAN        & Int8               & 6.32               & 80.22            & 47.41             & 88.9\%                 \\
        \bottomrule
    \end{tabular}
\end{table}

\subsection{Multi-Expert Scaling}
\label{sec:moe}

Single-head deployment is not the regime where SHARe-KAN's value compounds.
At one expert the MLP baseline is smaller (0.41 MB) and slightly more accurate (86.11\% VOC) than any SHARe-KAN variant; a practitioner deploying a single detection head should prefer the MLP.
SHARe-KAN's advantage emerges when a single backbone must serve many task-specific heads, where Dense KAN's per-expert memory cost becomes prohibitive.

\begin{figure}[t]
    \centering
    \includegraphics[width=0.7\linewidth]{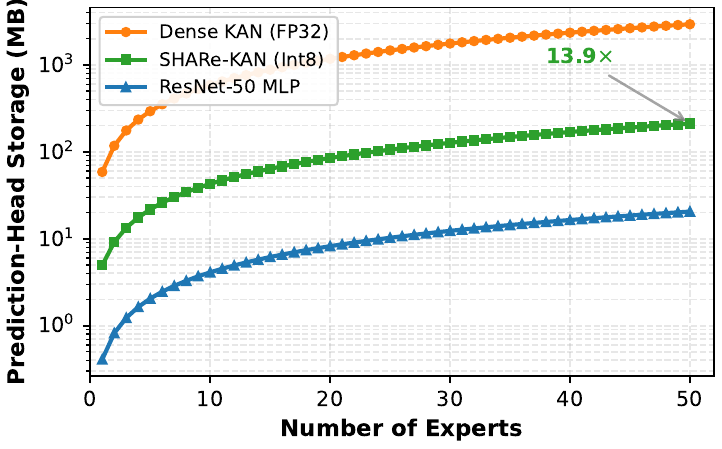}
    \caption{\textbf{Multi-expert memory scaling.} Prediction-head parameter storage as a function of expert count. SHARe-KAN Int8 stays within an order of magnitude of the MLP baseline while retaining KAN expressiveness; Dense KAN grows linearly to 2{,}933 MB at 50 experts.}
    \label{fig:moe_scaling}
\end{figure}

Figure~\ref{fig:moe_scaling} plots prediction-head parameter storage as a function of the number of experts, for the MLP, Dense KAN, and SHARe-KAN Int8.
At 50 experts the Dense KAN configuration requires 2{,}933 MB of prediction-head storage, which exceeds the on-chip memory of most edge accelerators and competes with the backbone for DRAM bandwidth.
The equivalent SHARe-KAN Int8 configuration requires 211 MB, a 13.9$\times$ reduction, and fits comfortably in the memory budgets of contemporary mobile SoCs.
The MLP remains smallest in absolute terms (20.5 MB at 50 experts) but does not offer the functional expressiveness of per-edge learned basis functions.

%% file: sections/06_conclusion.tex
\section{Discussion and Conclusion}

Pre-trained Vision KANs are functionally reducible but not structurally pruneable without a fine-tuning recovery loop.
SHARe-KAN replaces that loop with a one-shot post-training compiler: a Gain-Shape-Bias decomposition with a layer-shared codebook cuts prediction-head parameter storage by 9.3$\times$ on PASCAL VOC detection at a 2.0 point in-domain accuracy cost, and LUTHAM holds the resulting codebook in on-chip L2 during inference.
The single-head comparison against a ResNet-50 MLP baseline is clear: MLPs remain smaller and slightly more accurate, and we do not argue otherwise.
The regime where SHARe-KAN pays off is multi-expert deployment, where Dense KAN's per-expert storage grows linearly to 2{,}933 MB at 50 heads while SHARe-KAN Int8 stays at 211 MB.

\subsection{Limitations}

Evaluation is restricted to a single dataset (PASCAL VOC detection with zero-shot transfer to COCO) and a single backbone (ResNet-50).
We have not verified that the 9.3$\times$ compression ratio or the OOD retention behavior generalize to other vision tasks (segmentation, keypoint estimation, depth prediction), other backbones, or non-vision domains.
Quantization-induced OOD cost is small in this evaluation (1.3 retention points from FP32 to Int8), but the VQ clustering step itself costs roughly 10 retention points against the Dense KAN baseline; closing that gap is the main remaining accuracy cost of the approach and is open for future work.
All compression numbers report prediction-head parameter storage; inference-time peak memory is dominated by backbone activations ($\sim$4.3 GB for ResNet-50 at batch 1), so SHARe-KAN by itself does not reduce peak inference memory on this backbone.
Backbone compression is orthogonal and can be layered on top of SHARe-KAN, but we do not evaluate such a combination here.

\subsection{Future Directions and Broader Impact}

Two extensions follow from the multi-expert scaling result in \S\ref{sec:moe}.
First, a codebook pre-trained on a larger corpus (ImageNet, OpenImages) would reduce the VQ clustering cost on transfer and produce a visual dictionary that is fixed across tasks, a direction supported by recent work on shared parameter manifolds across disjoint training objectives~\citep{kaushik2025universal}.
Second, SHARe-KAN's low per-expert memory cost makes it a plausible primitive for edge MoE deployment, where dense expert weights currently dominate bandwidth and SHARe-KAN indices would instead fit inside a shared, cache-resident codebook.
On broader impact, DRAM access consumes 100 to 1000$\times$ more energy per bit than on-chip SRAM~\citep{horowitz2014energy}, so reducing DRAM traffic through cache residency lowers per-inference energy on bandwidth-constrained platforms; the same efficiency can also enable privacy-invasive surveillance, and appropriate regulatory frameworks are a separate question from the technical contribution.